%% file: ijcai26.tex
\documentclass{article}
\usepackage{ijcai26}

\usepackage{times}
\usepackage{soul}
\usepackage{url}
\usepackage[hidelinks]{hyperref}
\usepackage[utf8]{inputenc}
\usepackage[small]{caption}
\usepackage{graphicx}
\usepackage{amsmath}
\usepackage{amsthm}
\usepackage{booktabs}
\usepackage{algorithm}
\usepackage{algorithmic}
\usepackage[switch]{lineno}
\usepackage{amssymb}
\usepackage{mathrsfs} 

\title{Human-in-the-Loop Meta Bayesian Optimization for Fusion Energy and Scientific Applications}

\author{
    Ricardo Luna Gutierrez$^1$\and
    Sahand Ghorbanpour$^1$\and
    Rahman Ejaz$^2$\and
    Varchas Gopalaswamy$^2$\\
    Riccardo Betti$^2$\and
    Vineet Gundecha$^1$\and
    Aarne Lees$^2$\And
    Soumyendu Sarkar$^1$
    \thanks{Corresponding author.}
    \affiliations
    $^1$Hewlett Packard Enterprise\\
    $^2$University of Rochester\\
    \emails
    \{rluna, sahand.ghorbanpour, vineet.gundecha, soumyendu.sarkar\}@hpe.com,
    \{reja, vgop, betti, alees\}@lle.rochester.edu
}

\begin{document}

\maketitle

\begin{abstract}
    Inertial Confinement Fusion (ICF) holds transformative promise for sustainable, near-limitless clean energy, yet remains constrained by prohibitively high costs and limited experimental opportunities. This paper presents Human-in-the-Loop Meta Bayesian Optimization (HL-MBO), a framework that integrates expert knowledge with few-shot, uncertainty-aware machine learning to accelerate discovery in data-scarce, high-stakes scientific domains. HL-MBO introduces a meta-learned surrogate model with an expert-informed acquisition function to recommend candidate experiments. To foster trust and enable informed decisions, HL-MBO also provides interpretable explanations of its suggestions. We show that HL-MBO outperforms current BO methods for ICF energy yield optimization, as well as benchmarks in molecular optimization and critical-temperature maximization for superconducting materials. 
\end{abstract}

\begin{figure*}[!ht]
    \centering
    \includegraphics[width=1.0\textwidth]{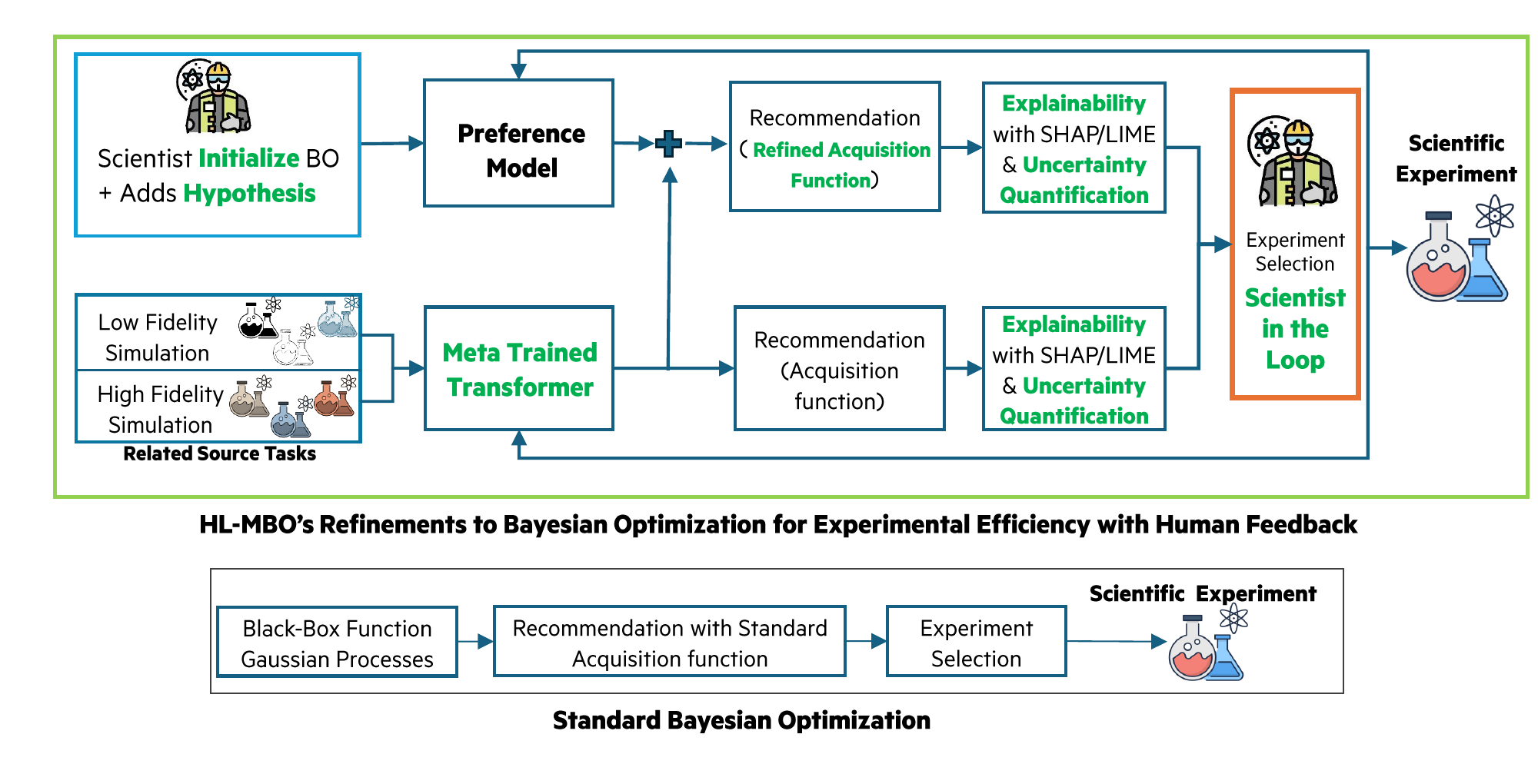}
    \vskip -0.3cm
    \caption{Refinements proposed in HL-MBO over standard Bayesian Optimization. HL-MBO trains a meta-surrogate using related source tasks and integrates a preference model to incorporate expert knowledge. During online black-box optimization the meta-surrogate and preference model are used in conjunction with an AF to optimize the target function (scientific experiment), providing explanations to support decision-making.}
    \label{figure:pipeline}
\end{figure*}

\section{Introduction}

The global community faces an existential challenge: the need to meet increasing energy demands while rapidly decarbonizing to mitigate the effects of climate change. Inertial Confinement Fusion (ICF) represents a transformative solution to this crisis, offering a path to a near-limitless, carbon-neutral power source without the long-lived radioactive waste or risks associated with traditional nuclear fission. 

However, ICF experimental facilities are among the most complex and expensive machines ever built. Consequently, experiments are extremely costly and  researchers are limited to only a few shots annually. For this reason, there is an urgent need for highly efficient optimization methods that can maximize the insights gained from every single experiment, effectively accelerating the timeline to a clean energy future.

Bayesian Optimization (BO) has shown promise in optimizing expensive black-box functions, using a surrogate model and acquisition function (AF) to guide experimental sampling~\cite{NIPS2015_11d0e628,NIPS2012_05311655,Wang2023}. However, BO faces skepticism due to its black-box nature and insufficient sample efficiency for complex tasks like ICF~\cite{Betti2016,PhysRevLett,Gopalaswamy2019-tc,Gopalaswamy2024-yn}.

Recent research incorporates human expertise into BO through human-in-the-loop methods, preference learning and explainability~\cite{Colella_2020,NEURIPS2022_6751611b,gupta2023bomuse,hvarfner2022pibo,adachi2024looping}. While promising, these methods are limited to single-task optimization, requiring the process to restart for each new task. To address new optimization tasks' sample efficiency and leverage accumulated knowledge, Meta-Learning within Bayesian Optimization (Meta-BO) has garnered increasing interest~\cite{bai2023transfer}. 

In Meta-BO, the surrogate model and/or AF are trained using Meta-Learning techniques on a set of source tasks, enabling rapid adaptation to new, unseen problems. Despite the promising transfer learning capabilities of Meta-BO, fundamental challenges remain unaddressed for expensive scientific experiments. The trustworthiness and explainability of Meta-BO decisions, necessary for adoption in high-stakes domains, have received little attention. Furthermore, the principled integration of valuable expert knowledge to potentially amplify optimization performance within Meta-BO remains an open problem. 

To address these limitations, we introduce \textbf{HL-MBO}, an explainable human-in-the-loop framework for Meta-Bayesian Optimization. Developed through a direct partnership between AI researchers and ICF physicists, HL-MBO integrates expert knowledge with few-shot, uncertainty-aware machine learning to accelerate discovery in data-scarce, high-stakes scientific domains. Built with interpretability in mind, HL-MBO provides insights into its decision-making process, helping to foster trust and expert engagement. An overview of the framework is presented in Figure~\ref{figure:pipeline}. We summarize our main contributions as follows: 

\begin{itemize}
\item We propose HL-MBO, the first framework that integrates human-in-the-loop and Meta-BO. We demonstrate how this framework achieves superior performance against state-of-the-art (SOTA) human-aided BO and Meta-BO methods in high-stakes applications such as ICF.
\item We propose a new formulation for an expert-informed AF, which integrates both preference learning and experts' hypotheses, and study the effect that human expertise can have over the optimization process.
\item We implement a comprehensive explainable framework for Meta-BO, leveraging uncertainty quantification, Shapley values and LIME to provide users with insights into the model's decision-making process and foster trust.
\end{itemize}

\section{Background}



\subsection{Challenges of Yield Optimization for ICF}

Inertial Confinement Fusion (ICF) achieves nuclear fusion by using lasers to compress and heat a deuterium-tritium fuel pellet to extreme temperatures and pressures, overcoming electrostatic repulsion between nuclei and releasing energy~\cite{Betti2016}. Energy yield is highly sensitive to the \textbf{laser pulse shape (LP)}, which drives ignition within 3 ns and is controlled by five critical parameters.

ICF experiments are extremely costly due to the sophisticated laser systems and harsh operating conditions. Moreover, access to ICF-capable facilities is limited, with only 5–10 shots typically conducted per day and just a few experimental days available each year.

Although simulators can generate data to train surrogate models or warm-start optimization, discrepancies between simulated and real outcomes often arise due to the system's complexity. As a result, there is a strong need for methods that can leverage simulation data but still be able to quickly adapt on the fly to compensate for these discrepancies.

\subsection{Meta Bayesian Optimization}

Bayesian Optimization (BO) is a technique that considers an objective function $f$ as a black-box and aims to find the global optimum by sequentially selecting points to evaluate based on the information obtained from previous evaluations~\cite{Garnett_2023}. Meta-Bayesian Optimization (Meta-BO) approaches aim to improve the optimization of new unseen target black-box functions by leveraging knowledge from a set of $N$ related source tasks (functions) $\mathcal{F}$~\cite{maraval2023endtoend}. We assume the model has access to this knowledge as datasets $\mathcal{D}_1, ...., \mathcal{D}_N$. Each dataset $\mathcal{D}_n$ consist of $e_n$ evaluations of $f_n(x) \in \mathcal{F}$ for all $n \in \{1, ..., N\}$, such as $\mathcal{D}_n = \{( x_n^i, y_n^i)\}^{e_n}_{i=1}$, where $ y_n^i = f_n(x_n^i)$.

In single-task scenarios, GPs have traditionally been the primary tool for constructing surrogate models. However, in the context of Meta-BO, recent studies highlight the superiority of uncertainty-aware meta-learning approaches, such as Neural Processes (NPs)~\cite{garnelo2018CNPs}. NPs are a class of models that combine the expressive function approximation capabilities of neural networks with the stochastic properties of GPs. Trained following a meta-learning framework, NPs enable rapid adaptation to new functions at test time~\cite{jha2023neural}.
Given a dataset of input-output context samples $\mathcal{D}_c$ and an unlabeled set of $l$ target locations $\mathcal{X}_T=\{x^{i}\}^{l}_{i=1}$ in which we would like to make predictions, NPs generate a distribution $p(\cdot | \mathcal{X}_T, \mathcal{D}_c)$ that approximates the true posterior over the target labels $\mathcal{Y}_T$.

\section{HL-MBO}
\label{headings}


HL-MBO is an explainable meta-learning human-in-the-loop framework that improves the efficiency and trustworthiness of black-box optimization. HL-MBO accelerates the optimization of high-stakes experiments by synthesizing historical task data with real-time expert insights, ensuring rapid convergence even in data-scarce environments.

The methodology presented here was developed through direct and iterative discussions between AI researchers and ICF physicists to ensure the framework addresses real-world experimental problems. Moreover, central to our approach is an expert-guided AF where the ICF experts acted as domain leads.

\subsection{Preference Learning}

Traditional Meta-BO approaches overlook the valuable insight that human experts can provide. In resource-intensive scientific applications such as ICF, effectively leveraging this expert knowledge is critical to increase experimental efficiency. To address this problem, HL-MBO incorporates a preference model built on binary preference learning~\cite{adachi2024looping,BRADLEY1952,chau2022}. We construct a preference dataset by presenting experts with pairs of candidate points, $\{x_1, x_2\}$, and asking them which point is more likely to be a better candidate for experimental evaluation. Specifically, experts label each pair based on their preferences. This process is repeated $M$ times, resulting in a dataset $D_\text{pref} = \{x^i_1, x^i_2, y_\text{pref}^i\}_{i=1}^{M}$, where $y_\text{pref}$ is $1$ if $x_1$ is favored over $x_2$, and 0 otherwise. $D_\text{pref}$ is then used to build a binary preference function $g$, which follows a likelihood model: $\mathbb{P}(y_\text{pref}, | x_1, x_2) = S(y_\text{pref}; g(x_1, x_2))$. Where $S(y_\text{pref};z) := z^{y_\text{pref}}(1-z)^{1-y_\text{pref}}$ represents a Bernoulli likelihood and $g(x_1,x_2)$ represents the user's preference level for $x_1$ compared to $x_2$. To model function $g$, we employ Dirichlet-based GPs~\cite{adachi2024looping,milios2018dirichletbased} combined with skew-symmetric data augmentation~\cite{pmlr-v151-lun-chau22a}. Following this approach allows us to estimate a preference model $\pi$ as GPs that generate a distribution $p_{\pi}(\cdot| \mathcal{X}_T, \mathcal{D}_c)$ that approximates an expert aligned posterior given a set of target locations $\mathcal{X}_T$ and context samples $\mathcal{D}_c$.

\subsection{Experts' Hypotheses} 
Typically, the set of pair candidate points $\{x_1, x_2\}$ presented to human experts is randomly selected~\cite{adachi2024looping}. However, this random process is impractical and faces resistance from human experts in complex tasks due to the large number of samples that require labeling. A more efficient approach would be to reduce the sampling space to high-quality zones and reduce the number of samples requiring labeling by improving sample quality. To achieve that, we explore the concept of experts' hypotheses ~\cite{cisse2024hypbo} in the scope of preference learning. Experts' hypotheses can be defined as subsets \( \mathcal{H} \subset \mathcal{X} \) of the overall search space, where experts identify a range of values considered promising. For instance, in a 2D problem (two features), a hypothesis might be specified as $\mathcal{H} = \left\{ \mathbf{x} \in \mathcal{X} \;\middle|\; 0.1 < \mathbf{x}_1 < 0.7,\; 0.5 < \mathbf{x}_2 < 0.9 \right\},$
where \( \mathbf{x}_1 \) and \( \mathbf{x}_2 \) refer to features 1 and 2, respectively. Instead of performing a selection of pairs across the entire search space, we constrain the selection to points that fall within these defined hypotheses. 
This constraint applies exclusively during $D_\text{pref}$ construction, without interfering with online black-box optimization's flexibility to explore beyond hypothesis boundaries. This minimizes influence on the optimization process while improving preference data collection efficiency and quality.

\subsection{Meta Training}

A key shortcoming of prior human-aided BO techniques is their inability to exploit the rich information contained within historical experimental data or simulations~\cite{gupta2023bomuse,hvarfner2022pibo,adachi2024looping,souza2021bayesian,ramachandran2020incorporating,chakraborty2024explainable,akata2020research,venkatesh2022human}. In this work, in parallel with constructing the preference model, we train a Meta-BO model using a set of related source tasks and TNPs. TNPs are particularly well suited for this setting as they provide a robust method for uncertainty-aware meta-learning by leveraging sequence modeling, enabling sample efficient optimization.
We consider each set $\mathcal{D}_n \in \{\mathcal{D}_1, ...., \mathcal{D}_N\}$ containing ${e_n}$ evaluations  as an ordered training sequence. A random subset of each training sequence is selected as the context set, comprising context pairs $(x_{1:m},y_{1:m})$, which serve as few-shot conditioning for the transformer model. Subsequently, the TNPs autoregressively model the predictive likelihood of the remaining $({e_n}-m)$ target points and optimize the objective:
\begin{equation}
\mathcal{L}(\theta) = \mathbb{E}_{x_{1:{e_n}}, y_{1:{e_n}}, m} \left[ \log p_{\theta}(y_{m+1:{e_n}} \mid x_{1:{e_n}}, y_{1:m}) \right].
\end{equation}

During each training iteration, the indices $n$ (dataset) and $m$ (context-target split) are picked uniformly to define the division of the samples into context and target points. For the transformer architecture in TNPs, we adopt the GPT-style model described in \shortcite{nguyen2023transformer}.

\subsection{Acquisition Function}

In expert-aided BO, we aim to find the point in the parameter space that leads to the best performance while considering experts' preferences. CoExBO~\cite{adachi2024looping} established a principled approach by combining GP-based surrogate model $\mathcal{S}$ and preference model $\pi$ through a UCB acquisition function $\alpha_{\mathcal{S},\pi}$ as:

\begin{align}
    \alpha_{\mathcal{S},\pi}(x) &:= \mu_{\mathcal{S},\pi}(x) + \sigma_{\mathcal{S},\pi}(x) \\ 
    \mu_{\mathcal{S},\pi}(x) &:= \frac{\sigma^2_{\mathcal{S},\pi}(x)}{\mathscr{S}^2_\pi(x)}\mu_{\pi}(x) + 
    \frac{\sigma^2_{\mathcal{S},\pi}(x)}{\sigma^2_\mathcal{S}(x)}\mu_{\mathcal{S}}(x) \label{equation:2} \\ 
    \sigma^2_{\mathcal{S},\pi}(x) &:= \frac{\mathscr{S}^2_\pi(x)\sigma^2_\mathcal{S}(x)}{\mathscr{S}^2_\pi(x)+\sigma^2_\mathcal{S}(x)} \label{equation:3} \\ 
    \mathscr{S}^2_\pi(x) &:= \sigma^2_\pi(x) + \gamma{t^2}\sigma^2_S(x), \label{equation:5}
\end{align}

where $\mu_{\mathcal{S}}, \sigma^2_{\mathcal{S}}$ and $\mu_\pi,\sigma^2_\pi$ are the predicted mean and variance (uncertainty) of the surrogate model $\mathcal{S}$ and the preference model $\pi$, respectively. The parameter $\gamma$ is a decay factor and $t$ represents the current time step in the optimization process. $\gamma$ ensures that the information provided by $\pi$ decays, so the impact of $\pi$ is high at the start of the optimization but allows $\mathcal{S}$ to take over at later stages.  As demonstrated by \cite{adachi2024looping}, this formulation offers a no-harm guarantee, where even in a worst-case scenario the preference-influenced AF achieves convergence rates comparable to or better than those of standard BO AFs.

While CoExBO demonstrated strong single-task performance, it cannot leverage knowledge transfer from prior experiments or simulations, a crucial limitation for expensive scientific optimization where historical data from related tasks is available. Additionally, empirical studies have shown Expected Improvement (EI) often outperforms UCB in practical optimization scenarios~\cite{JMLR:v22:18-220}.

To address these limitations, we introduce a meta-learned TNP surrogate $\mathcal{S}$ to replace the standard GP-based surrogate used in traditional approaches. This enables knowledge transfer from related source tasks. Second, we formulate a new human-aided AF based on Expected Improvement to better exploit the improvement-based criterion. Our EI-based formulation $\alpha_{\mathcal{S},\pi}(x)$ is defined as:

\begin{align}
\alpha_{\mathcal{S},\pi}(x) &= (\mu_{\mathcal{S},\pi}(x) - f(x^+) - \xi) \Phi(Z) + \sigma_{\mathcal{S},\pi}(x) \phi(Z)\\
Z &= \frac{\mu_{\mathcal{S},\pi}(x) - f(x^+) - \xi}{\sigma_{\mathcal{S},\pi}(x)},
\end{align}

where $f(x^+)$ is the best observed value so far, $\Phi(\cdot)$ is the cumulative distribution function (CDF) and $\phi(\cdot)$ is the probability density function (PDF) of $Z$~\cite{TACQ2010467}. The parameter $\xi$ is a small positive hyperparameter that encourages exploration. Crucially, we retain the decay-based uncertainty combination structure from Equations~\ref{equation:2}-\ref{equation:3} (see Technical Appendix C), which preserves the intended no-harm behavior of progressively reducing preference influence over time. We then replace the underlying components $\mu_\mathcal{S}$ and $\sigma^2_\mathcal{S}$ with the predictive mean and variance derived from our meta-trained TNP, enabling meta-learning capabilities within the same acquisition framework.

To provide users with flexible optimization strategies, HL-MBO also offers a preference-free AF $\alpha_\mathcal{S}(x)$ that applies standard Expected Improvement over the meta-surrogate alone ($\mu_\mathcal{S}$ and $\sigma^2_\mathcal{S}$), enabling pure data-driven optimization.

\subsection{Explainability}
\label{subsection:explainability}

\begin{figure*}[h]
    \centering
    \includegraphics[width=1.0\textwidth]{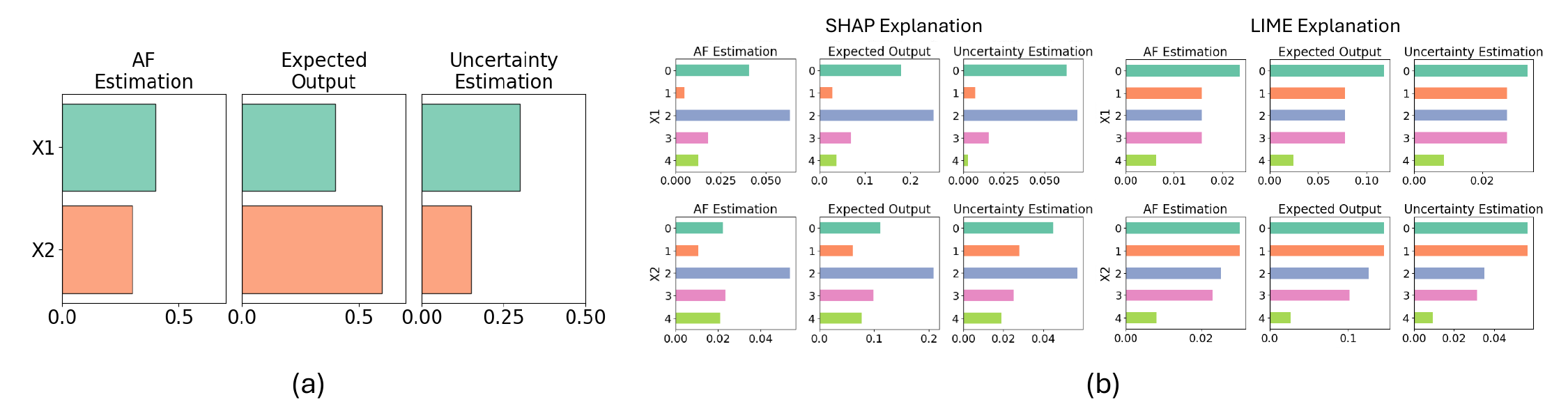}
    \caption{(a) Visual representation of the expected experimental output (TNPs' Mean) and the associated uncertainty of the prediction for both candidate points proposed. (b) Feature attribution using SHAP and LIME on the ICF optimization task over 5 features, for the two candidate points proposed by HL-MBO. Each bar represents contribution with respect to the BO metrics (AF, Expected Output, Uncertainty Estimation).}
    \label{figure:shap}
\end{figure*}

To assist human experts in making informed decisions during the optimization process, we provide explanations of the model predictions. First, we present visual representations of the TNPs' predicted experimental output (TNPs Mean) and its uncertainty levels, along with the AF values for the proposed HL-MBO candidates. Presenting uncertainty is particularly important because it highlights the confidence level of the model in its predictions. Figure \ref{figure:shap}(a) shows an example of these representations for general applications. We include representations for the specific task of ICF energy yield optimization in Section B of the Technical Appendix.

Moreover,  we incorporate two widely used explainability methods: Shapley values (SHAP)~\cite{P-295,NIPS2017_7062} and Local Interpretable Model-Agnostic Explanations (LIME)~\cite{rodemann2024explaining,ribeiro2016whyitrustyou}.

Figure \ref{figure:shap} (b) provides an example of the explanation generated by SHAP and LIME for the user. "X1" and "X2" denote the two candidate points proposed by $\alpha_\mathcal{S}$ and $\alpha_{\mathcal{S}, \pi}$, respectively. For each candidate point, the bars represent the Shapley or LIME attribution value of the features in relation to their respective AF score, TNPs' predicted mean and TNPs' uncertainty at proposed location. The divergence in feature attributions across methods offers human experts complementary perspectives on the model's reasoning. Providing these diverse explanations empowers experts to identify truly influential features with greater confidence, facilitating precise and targeted interventions in the optimization process. Furthermore, this multi-faceted explainability enables experts to critically analyze and potentially discard candidate solutions when the observed feature importance deviates substantially from their domain knowledge or expected behavior, fostering a more robust and reliable optimization loop. In our collaboration, ICF experts utilized these attributions to verify if suggested laser pulse parameters aligned with known physical sensitivities, allowing them to confidently override or accept model-predicted "shots" based on domain intuition.

\subsection{Explainable Expert Guided Optimization}

Given a trained meta-surrogate model $\mathcal{S}$, a preference model $\pi$, a target function $f_\text{target}$, and a context dataset $D_c$ which contains a set of $\mathcal{I}$ initial input-output samples which have been evaluated on $f_\text{target}$, we start the optimization process by generating a pair of candidate points $x_1$ and $x_2$ where:
\begin{equation*}
x_1 = \mathop{\text{argmax}}_{x \in X}\alpha_\mathcal{S}(x) \quad \quad x_2 = \mathop{\text{argmax}}_{x \in X}\alpha_{\mathcal{S}, \pi}(x)
\end{equation*}
Given these candidates, we provide explanations about them and the expert selects between both options. Subsequently, this newly selected input-output pair $\{x_i, y_i\}$ is appended to the dataset $D_c$, $D_c \leftarrow D_c \cup \{x_i, y_i\}$. $D_c$ is used as context for the surrogate to adapt its predicted posterior over the current function $f_\text{target}$ and a new pair of candidate points is generated. Since we are using TNPs, this adaptation does not require any gradient updates and only a forward pass with the new context $D_c$ as input is necessary. This iterative process continues until a budget of points to evaluate $B$ is reached. HL-MBO's overall optimization process is shown in Algorithm \ref{algo:algo1}.

\begin{algorithm}[h]
\textbf{Input:} Meta-surrogate $\mathcal{S}$, preference model $\pi$, target function $f_\text{target}$, initial samples $\mathcal{I}$, budget $B$. \\
\begin{algorithmic}[1]
\STATE Initialize $D_c \leftarrow \mathcal{I}$ input-output pairs.
\FOR{$i = 1$ to $B$}
    \STATE Obtain posterior $p(\cdot| D_c)$ from $\mathcal{S}$ and $\pi$.
    \STATE Propose candidates $\{x_1,x_2\}$ using $\alpha_\mathcal{S}$ and $\alpha_{\mathcal{S}, \pi}$.
    \STATE Present candidates with explanations to expert.
    \STATE Expert selects $x_p \in \{x_1,x_2\}$, evaluate $f_\text{target}(x_p) = y_p$.
    \STATE Update $D_c \leftarrow D_c \cup \{x_p, y_p\}$.
\ENDFOR
\end{algorithmic}
\caption{HL-MBO algorithm}
\label{algo:algo1}
\end{algorithm}

\section{Experiments}
\label{sec:experiments}
We evaluate HL-MBO across diverse domains to address three research questions: \textbf{(1)} performance on optimizing black-box tasks, \textbf{(2)} the influence of expert preferences, and \textbf{(3)} the impact of experts' hypotheses.

For \textbf{ICF}, we evaluate HL-MBO in the task of energy yield optimization, where we optimize 5 critical parameters of a LP in order to maximize energy generation.

Moreover, to demonstrate generalization across domains we make use of the HPO-B benchmark~\cite{HPO-B} a popular hyperparameter optimization benchmark used to evaluate Meta-BO approaches. We selected 3 different datasets for this evaluation; Ranger (6D), SVM (8D), and XGBoost (16D), covering problems from 6 to 16 input dimensions (hyperparameters to optimize).

Additionally, we evaluate HL-MBO in the real-world high dimensional tasks of critical temperature maximization for \textbf{Superconductor} (86D) materials and practical molecular optimization \textbf{(PMO)} (2048D)~\cite{gao2022sample,designbench}.

Additional experimental details and dataset descriptions can be found in the Technical Appendix.
 
\subsection{HL-MBO Performance}
\label{sec:performance}
To answer the first question, we compare HL-MBO against CoExBO and $\pi$BO~\cite{hvarfner2022pibo,adachi2024looping}, SOTA approaches for human-in-the-loop BO, NAP~\cite{maraval2023endtoend}, the SOTA for Meta-BO, and a standard TNP+EI~\cite{nguyen2023transformer}. Moreover, to evaluate the isolated impact of our proposed AF, we design a new baseline, a meta-learning version of CoExBO (MCoExBO). MCoExBO uses the same \textbf{UCB-based} AF proposed in CoExBO; however, the single-task GPs are replaced by TNPs. Both HL-MBO and MCoExBO were meta-trained in the same training sets. Similarly, for fairness of comparison, the GPs used by CoExBO were initialized with training data. For all cases, the methods were evaluated on a set of test tasks that have not been seen during training.

For \textbf{ICF}, the ICF experts actively formulated hypotheses and selected from the candidate experiment pairs. Two ICF physicists provided 100 pairwise labels in about one hour, and two ICF experts selected between the two candidates during online optimization. For the \textbf{HPO-B, superconductor and PMO} tasks, we simulate human decision making for selecting candidate points proposed by the AFs $\alpha_{\mathcal{S}}$, $\alpha_{\mathcal{S}, \pi}$ and constructing $D_{pref}$, by employing the synthetic human selection method outlined in \shortcite{adachi2024looping}. In this simulation, user selection is modeled as $f_{human}(x_1)$ \textgreater $f_{human}(x_2)$, where $f_{human}(x) := f_{target}(x) + \epsilon_\text{pref}$ and $\epsilon_\text{pref} \sim \mathcal{N}(x;0,\sigma^2_\text{pref})$. For hypothesis selection, we simulate expert knowledge by dividing the search space into slices of size $M=100$ and choosing the slice that contains the optima. For our experiments $\sigma^2_\text{pref} = 0.1$.

\begin{figure}[h]
    \centering
    \includegraphics[width=1.0\columnwidth]{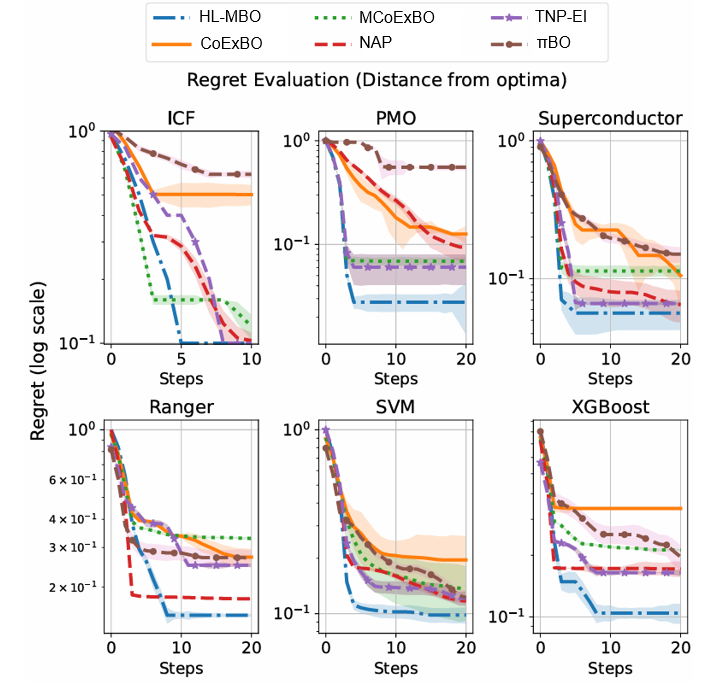}
    \caption{Results of the comparison between HL-MBO and the baselines. The lower the better. The shaded areas represent a $\pm1$ standard deviations. HL-MBO outperforms the baselines on all 6 benchmarks. Moreover, we can observe that HL-MBO is able to achieve the best sample efficiency in the scientific applications of ICF, PMO and Superconductor.}
    \label{figure:regret_evaluation}
\end{figure}

We set $\mathcal{I} = 1$ for all tasks to emulate a real-world setting where sampling is costly. We evaluated all methods using 25 random seeds for the HPO-B and superconductor tasks and 5 random seeds for the ICF and PMO tasks. 

As is common in BO literature, we evaluate performance in terms of simple regret, defined as $R_t = f(x^*) - f(x^+_t)$, where $x^+_t$ is the best input point found up to and including time step $t$~\cite{volpp2020maf,maraval2023endtoend}.  Given that, on average, a maximum of 10 shots can be fired during an ICF experiment day, we set our optimization budget $B=10$ for the ICF evaluations. Figure \ref{figure:regret_evaluation} presents the results of this evaluation. HL-MBO significantly outperforms all baselines, achieving the lowest regret with the fewest optimization steps. For ICF, HL-MBO reaches the optimum in 4 samples compared to 9 for the other methods, achieving a 44\% improvement in sample efficiency. 

As shown by MCoExBO, replacing the GP-based surrogate with TNPs improves optimization performance. However, using TNPs alone is insufficient, as MCoExBO does not match the performance of HL-MBO. This highlights the effectiveness of our AF formulation in driving superior optimization outcomes.

A more granular analysis of HL-MBO’s success in the ICF domain is provided in Technical Appendix B.

\subsection{Robustness Analysis}
\label{sec:hypthesis}

\paragraph{Hypothesis Evaluation} In this ablation, we explore the influence of experts' hypotheses on the performance of HL-MBO. For this experiment, we assume access to the target function $f_{target}$ and consider three experimental settings: Expert Hypothesis (EH), Random Hypothesis (RH), and Adversarial Hypothesis (AH). To construct these hypotheses, we partition the search space $\mathcal{X}_T$ in $K$ slices of size $M$. For each slice $k \in K$, we evaluate all input values $x_k^i \in \mathcal{X}_T^k$ and generate datasets $D_k = \{(x_{k}^i, y_{k}^i)\}^M_{i=1}$. 
The Expert Hypothesis represents an optimal scenario. We construct it by selecting the slices that contain the optima.
On the other hand, the Adversarial Hypothesis simulates the worst-case scenario, we choose the slice $k$ where the cumulative sum $V_k=\mathop{\sum}_{i=0}^My_k^i$ is the lowest. For the Random Hypothesis, we uniformly sample from $\mathcal{X}_T$. We also added CoExBO and NAP for comparison. 

\begin{figure}[h]
    \centering
    \includegraphics[width=0.85\columnwidth]{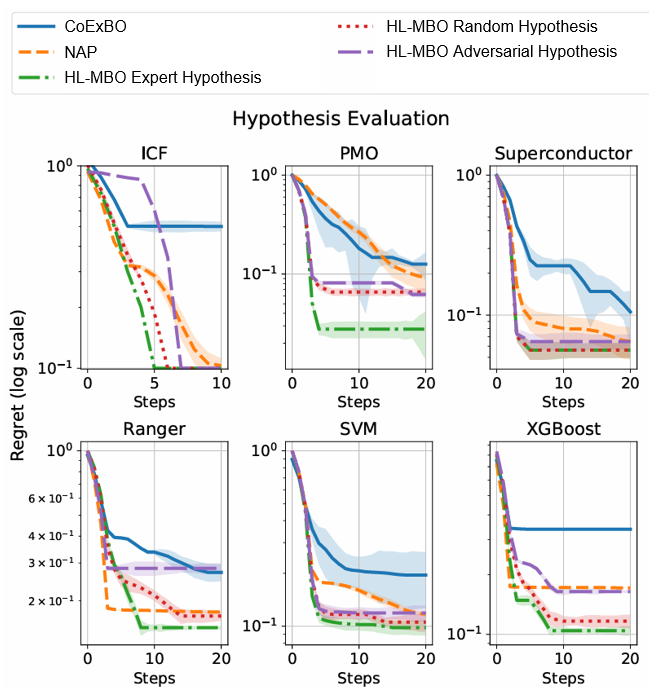}
    \caption{Results of the comparison between different hypothesis. Expert and Adversarial represent the best-case and worst-case scenarios, respectively, for sampling in the preference learning initialization. Random, represents uniform sampling.}
    \label{figure:ablation_evaluation}
\end{figure}

Figure \ref{figure:ablation_evaluation} presents the results of this ablation study. Despite the Adversarial Hypothesis impacting HL-MBO's performance, it still outperforms CoExBO and matches NAP's performance in most tasks, highlighting HL-MBO's robustness. Moreover, the Random Hypothesis surpasses NAP on average across the six different tasks, demonstrating the HL-MBO's ability to maintain strong performance even without access to an optimal hypothesis.

\begin{figure}[h]
    \centering
    \includegraphics[width=0.85\columnwidth]{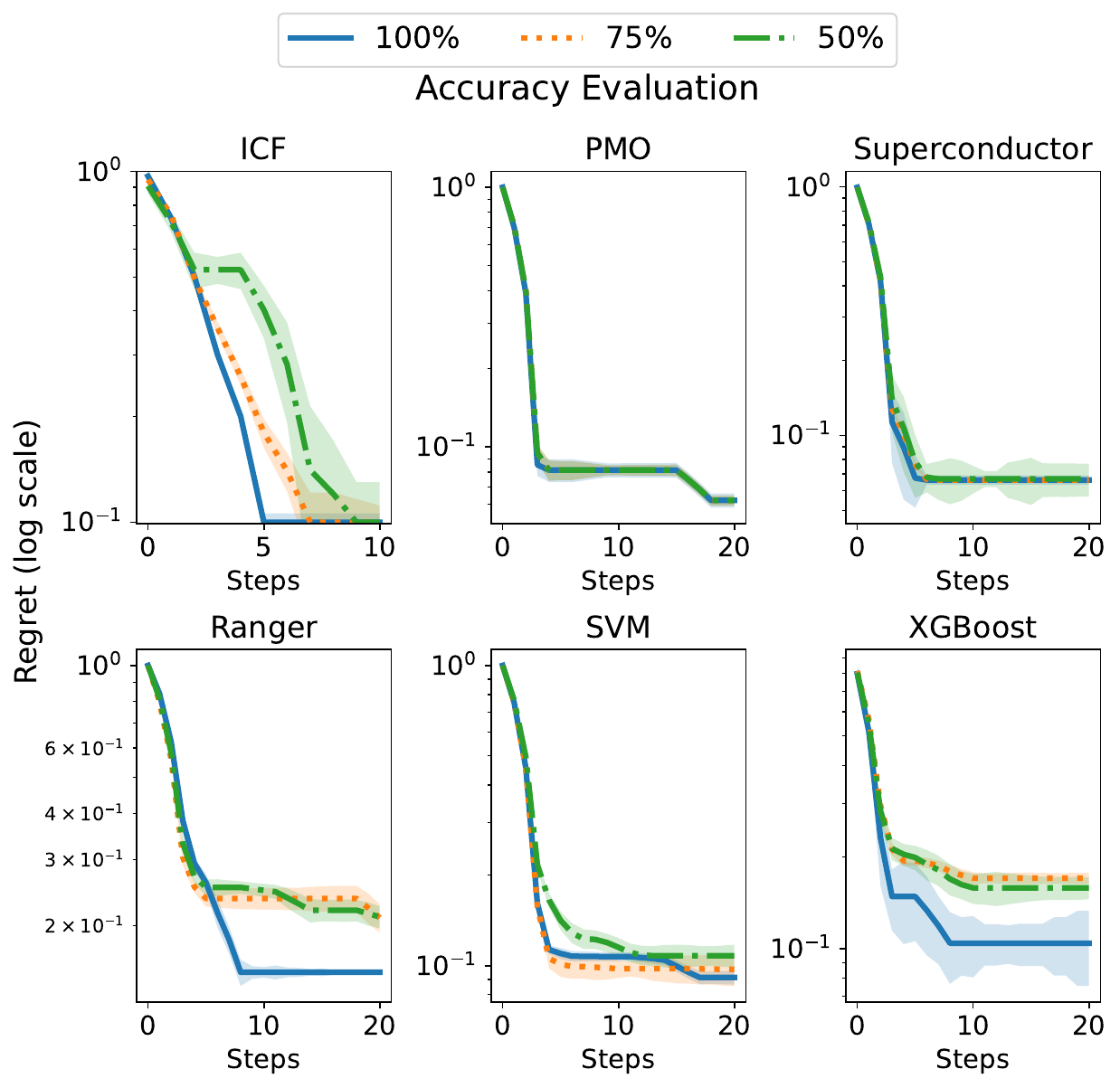}
    \caption{Results showing the accuracy of experts in selecting the better point from each presented pair during preference model construction. 50\% accuracy indicates random choice.}
    
    \label{figure:accurary_evaluation}
\end{figure}

\begin{figure}[h]
    \centering
    \includegraphics[width=0.85\columnwidth]{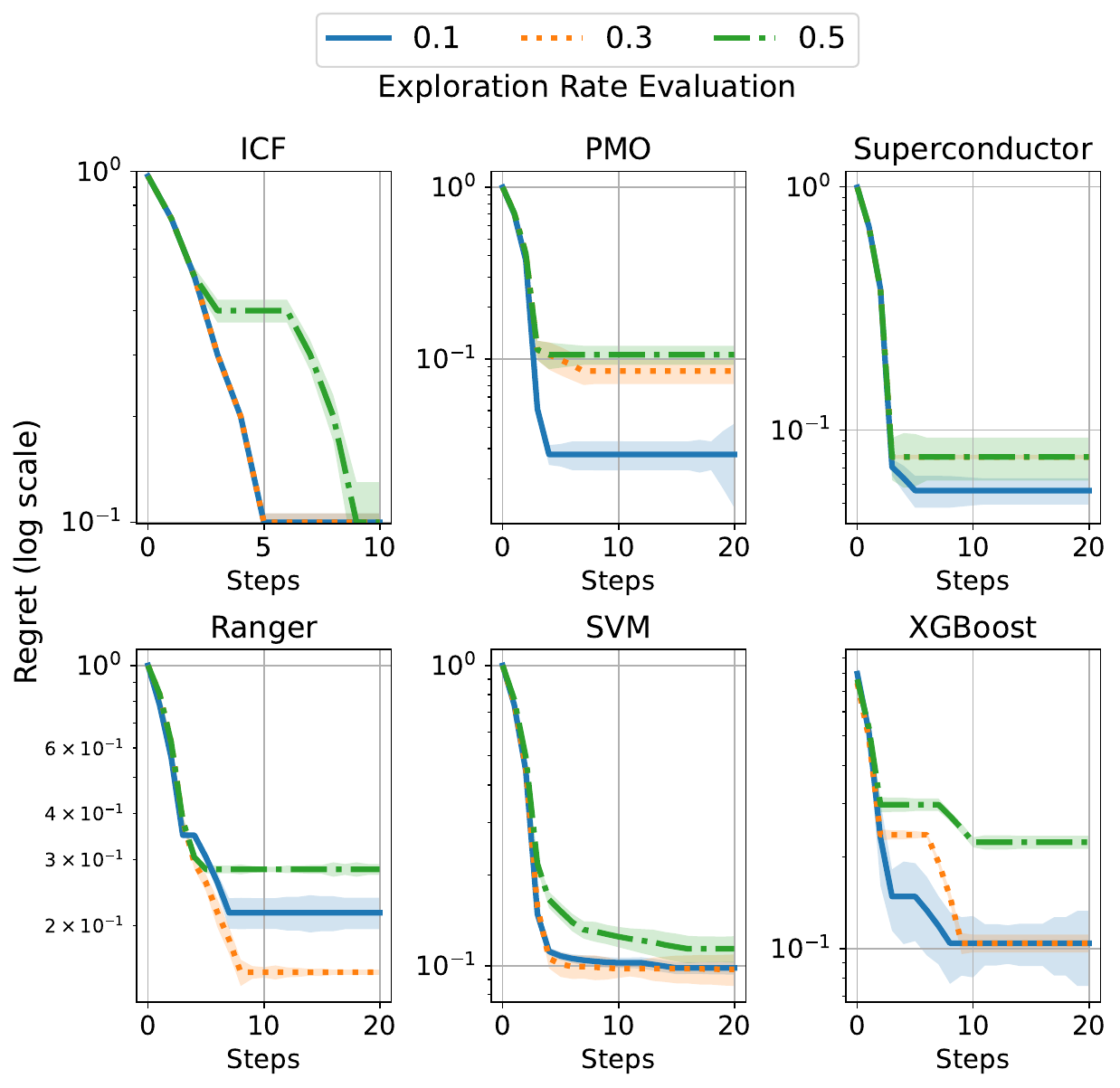}
    \caption{Results of the comparison between different values of $\xi$, which controls the level of exploration in HL-MBO's AF.}

    \label{figure:exploration_evaluation}
\end{figure}


\label{sec:accuracy}

\paragraph{Accuracy Evaluation} We study the effect of correctness in experts' selections when building dataset $D_\text{pref}$. Specifically, we assess how the accuracy in selecting the input value $x_p$ that yields the highest output $y_p$ from the candidate pairs $\{x_1, x_2\}$ affects HL-MBO's performance. We evaluate the performance of HL-MBO on different experts' accuracy thresholds: 50\%, 75\% and 100\%.  To conduct this analysis, we assume access to the target function, which allows us to evaluate all pairs of inputs in $D_\text{pref}$ and generate the corresponding datasets for each accuracy level. For all three cases we consider a good hypothesis. Figure \ref{figure:accurary_evaluation} shows the results of this evaluation. In most cases, $75\%$ and $100\%$ accuracy levels, which represent some level of knowledge about the black-box function by the expert, show better performance than random selection ($50\%$), emphasizing the crucial role of expert input in the optimization process. Moreover, we can see that in most cases even $75\%$ offers significant benefits.

\paragraph{Exploration Hyperparameter Evaluation} We investigate the influence of the parameter $\xi$ which governs the balance between exploration and exploitation in our EI-based AF. Higher values of $\xi$ encourage more exploration. The results of this evaluation are shown in Figure \ref{figure:exploration_evaluation}. Across all domains, smaller values of $\xi$ within the range of $0.1$ and $0.3$ generally yield better performance compared to $0.5$. However, domains such as Ranger and SVM require a higher level of exploration, as $\xi = 0.3$ obtains the best performance. These findings highlight that the optimal value of $\xi$ is domain-specific and must be carefully tuned to suit the problem at hand. Following standard practices, we use validation sets to fine-tune $\xi$ for our experiments.

\section{Related Work}

\paragraph{Human-aided and Explainable BO} 
While no prior work has integrated few-shot meta-learning into human-aided explainable BO as proposed in this paper, recent efforts have advanced both explainable and human-in-the-loop BO.

One methodology is to incorporate human knowledge into BO by encoding it as a prior over the input space~\cite{hvarfner2022pibo,souza2021bayesian,ramachandran2020incorporating,cisse2024hypbo,hvarfner2024a}, guiding the search toward promising regions. However, this requires users to precisely express their expertise, which becomes challenging in high-dimensional spaces~\cite{mikkola2021prior}.

Another strategy engages humans during optimization, allowing them to reject or label points based on intuition or expertise~\cite{gupta2023bomuse,adachi2024looping,chakraborty2024explainable,akata2020research,venkatesh2022human,borji2013bayesian}. This interactive strategy enables more adaptive exploration of the search space and presents the expert with a simpler avenue to integrate their knowledge. 

Methods for explainable BO have also emerged, providing insight into BO decisions to improve user understanding~\cite{adachi2024looping,rodemann2024explaining,chakraborty2023post,li2020explainability}. These approaches aim to improve user understanding of the decision-making process, providing insight during or after optimization to support a more informed evaluation of candidate solutions.

\paragraph{Optimization for ICF} Standard BO has been applied to target design and experimental optimization in ICF~\cite{chung2020offlinecontextualbayesianoptimization,BOBatch2024,Hatfield_2019,li2023hybrid,10.1063/5.0129565}. Additionally, surrogate models built from ICF simulations or historical data have been used for efficient experimental optimization~\cite{chung2020offlinecontextualbayesianoptimization,Humbird_2022,Wu_Yang_2022,10.1063/5.0215962,10.1063/5.0228824,Wal_McClarren_Humbird_2023}. However, these approaches have yet to explore meta-learning or human-aided strategies to enhance trustworthiness and performance.

\section{Conclusion}
\label{sec:conclusion}
We presented HL-MBO, an explainable human-in-the-loop Meta Bayesian Optimization approach. We evaluated our method across three different hyperparameter optimization tasks and the real-world tasks of energy yield optimization in ICF, practical molecular optimization, and critical temperature maximization for superconductor materials.  Our results showed that HL-MBO surpasses current SOTA for human-AI collaboration and Meta-BO. Moreover, we assessed the impact of experts' knowledge on Meta-BO and showed that its integration is beneficial. 

Our work can have a significant impact on real-world scientific applications where experiments are costly but expert knowledge is available. In a critical domain like ICF, we demonstrate that HL-MBO outperforms all baselines. Advancing research in this direction, brings us closer to the goal of developing nearly limitless clean energy, which could have profound environmental and societal benefits.

\subsection{Limitations} 
Meta-BO requires access to a set of related source tasks which might not be available in some cases. Moreover, training TNPs can be computationally expensive. While effective, eliciting precise and comprehensive hypotheses from domain experts in complex, high-dimensional real-world settings is challenging, suggesting avenues for future research into interactive or automated elicitation methods. Furthermore, our method requires input from human experts, which might not be available in all cases.

\section*{Acknowledgments} This work was supported by the U.S. Department of Energy, Office of Science, under Award No. DE-SC0024408 and DE-SC0024381.



\newpage
\input{appendix.tex}

\newpage
\bibliographystyle{named}
\bibliography{ref}

\end{document}

%% file: appendix.tex





\urlstyle{same}







\pdfinfo{
/TemplateVersion (IJCAI.2026.0)
}

\appendix

\twocolumn[
\begin{center}
    {\LARGE \textbf{Technical Appendix}}
\end{center}
\vspace{1em} 
]

\section{Experimental Details}

For HL-MBO and MCoExBO the autoregressive version of TNPs (TNPs-A) \cite{nguyen2023transformer} is used for all experiments. We follow the hyperparameters recommended in the original paper, as presented in Table \ref{table:hyperparameters}. For the learning rate, a Cosine annealing scheduler was used. We used Adam as optimizer \cite{kingma2017adammethodstochasticoptimization}. For selecting the value of $\xi$, we evaluated the approach using a validation set in small set of values $\xi \in \{0.1, 0.3, 0.5\}$ and selected the best performer as shown in Table \ref{table:mbohfhyperparameters}. For seeding, we used the native "random" python library.

For CoExBO and MCoExBO, the proposed values in the original CoExBO work for the exploration-exploitation weight, were used \cite{adachi2024looping}. For HL-MBO, CoExBO and MCoExBO a value of $\gamma = 0.1$. CoExBO code can be found in \url{https://github.com/ma921/CoExBO}.

For $\pi$BO, we constructed the required priors for each task by computing the mean and standard deviation of the optima across all training tasks. $\pi$BO code can be found in \url{https://github.com/piboauthors/PiBO-Hypermapper}.

 Computation for training and evaluation was performed on a server with a AMD EPYC 7542 32-Core CPU and two H-100 GPUs for 380 hours at around 60\% capacity, excluding $\pi$BO. For $\pi$BO, evaluations required around 510 hours of CPU computing, using the code provided by the authors. This might be due to being GPs and CPU based which struggled with the high-dimensional problems.
 
\subsection{Expert Involvement and Runtime}

The framework was developed through direct collaboration between AI researchers ($n=4$), who led model design and implementation, and ICF physicists ($n=4$), who defined the experimental requirements, provided the ICF datasets, and iteratively validated the workflow. For the ICF case study, two ICF physicists provided 100 pairwise preference labels for constructing $D_{\text{pref}}$, completing the labeling process in approximately one hour. During online optimization, two ICF experts selected between the candidate experiments proposed at each iteration.

The meta-surrogate training time was approximately 7 hours for PMO (2048D) and 3 hours for Ranger (6D) on a single H100 GPU. Once trained, TNP inference required less than 1 second, fitting the preference GP on 100 samples required less than 1 minute, and the end-to-end optimization step including explanation generation takes approximately 20 seconds.

\begin{table}[h!]
      \centering
      \small
      \begin{tabular}{|ll|}
        \hline
        \multicolumn{2}{|c|}{TNPs' Hyperparameters} \\ \hline
        \multicolumn{1}{|l|}{Model dimension}       & 64      \\ \hline
        \multicolumn{1}{|l|}{Embedding layers}      & 4       \\ \hline
        \multicolumn{1}{|l|}{Feed forward dimension}& 128     \\ \hline
        \multicolumn{1}{|l|}{Attention heads}       & 8       \\ \hline
        \multicolumn{1}{|l|}{Transformer layers}    & 6       \\ \hline
        \multicolumn{1}{|l|}{Dropout}               & 0.0     \\ \hline
        \multicolumn{1}{|l|}{Learning rate}         & $5e-5$ \\ \hline
      \end{tabular}
      \caption{Hyperparameters used by the TNPs for all experiments.}
      \label{table:hyperparameters}
\end{table}

\begin{table}[h!]
    \centering
    \small
    \begin{tabular}{|l|r|}
        \hline
        \multicolumn{2}{|c|}{HL-MBO AF Hyperparameters} \\ \hline
        \multicolumn{1}{|c|}{Task} & \multicolumn{1}{c|}{$\xi$} \\ \hline
        \multicolumn{1}{|l|}{ICF}     & 0.1 \\ \hline
        \multicolumn{1}{|l|}{PMO}   & 0.1 \\ \hline
        \multicolumn{1}{|l|}{Superconductor}    & 0.1 \\ \hline
        \multicolumn{1}{|l|}{Ranger}      & 0.3 \\ \hline
        \multicolumn{1}{|l|}{SVM}  & 0.3 \\ \hline
        \multicolumn{1}{|l|}{XGBoost} & 0.1 \\ \hline
    \end{tabular}
    \caption{Hyperparameter $\xi$ used for each task in for HL-MBO's evaluation.}
    \label{table:mbohfhyperparameters}
\end{table}

\subsection{Task Datasets}

For the \textbf{HPO-B} tasks we used the Meta-Dataset for XGBoost, SVM and Ranger provided in \url{https://github.com/machinelearningnuremberg/HPO-B}. We used the training, validation and test tasks were used in totality as defined the benchmark in our experiments.

To create the \textbf{ICF dataset}, the LOTUS library was used to generate laser pulse shapes based on an ad-hoc parametrization \cite{osti_2356817}. Laser power and time values derived from this parametrization served as inputs for LILAC \cite{PhysRevA.36.3926}, a physics-based laser fusion simulator. For a given fusion fuel target, the laser pulse shape determines experimental outcomes (e.g., neutron yield), affecting both the compression of the target and the growth of hydrodynamic instabilities \cite{10.1063/5.0069372}.
To construct the response surface, 5 laser parameters from the front end of the pulse shape representation were varied, generating 50k samples using Latin hypercube sampling while respecting the laser system’s design constraints. These laser pulse shapes were then used in LILAC simulations with a fixed fusion fuel target. Post-processing of simulation results provided neutron yield values, which were used to build the response surface with respect to the 5 laser parameters.
Two source tasks, one validation and one test task were created by altering the physics models in the simulator, specifically the equation of state, which impacts shock propagation in the fusion fuel materials and changes the response surface. This perturbation to the physics model allowed us to simulated the divergence between simulation and real experiments, giving us a pathway to evaluate the method in a realistic manner.

For \textbf{PMO} and \textbf{Superconductor}, we use datasets from \cite{gao2022sample} and \cite{designbench}, respectively. To obtain the PMO data we used the tdc python library \cite{Huang2021tdc} and \textbf{osimertinib} as an oracle. For the \textbf{Superconductor} tasks, we use the dataset as provided in \url{https://github.com/brandontrabucco/design-bench/tree/new-api}. To create distinct source, validation and test tasks, we partitioned the datasets by sub-sampling them, generating 6 training tasks, 3 test tasks, and 2 validation tasks.  We ensured that there was no overlap between the samples used in the source tasks, validation and test tasks. By separating the data in this way, we simulate conditions where the model must generalize to new, unseen tasks during evaluation while being trained only on related but distinct source tasks.

For HPO-B, PMO and Superconductor, the synthetic human preference mechanism is used as a controlled evaluation protocol to compare methods under standardized noisy pairwise feedback following prior human-in-the-loop BO work \cite{adachi2024looping}. These experiments are intended to measure optimization behavior under a consistent preference model rather than to reproduce the full richness of real expert judgment.

\section{Understanding HL-MBO's Performance in ICF}

For ICF the model predictions, uncertainties and AF can be better visualize using a 2D heatmap, where the search space can be sliced and presented in parameter pairs. In this case, we present the "Energy Yield" as the intensity in the heatmap over two of the most influential laser pulse shape optimization parameters "Foot Power" and "Picket Power. This gives the ICF scientist a clearer understanding of the surrogate insights and lets them better understand why those experiments are being proposed by the AFs. Moreover, this allows us to better analyze the performance of the models as shown in the next subsections.

\begin{figure*}[ht]
    \centering
    \includegraphics[width=1\textwidth]{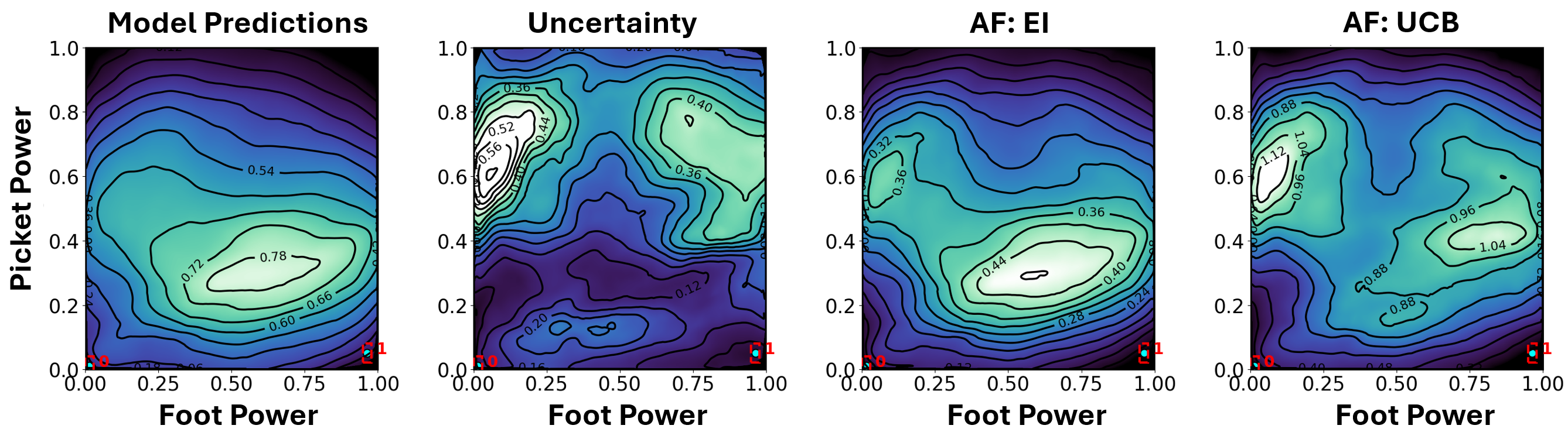}
    \caption{Illustration of an surrogate predictions, uncertainty and AFs value. Such visualizations provide ICF practitioners with insights to select the candidate most likely to succeed. The numbering represent the order in which samples were taken.}
    \label{figure:icf_representations}
\end{figure*}

\subsection{HL-MBO's Surrogate Predictions}

\begin{figure}[ht]
    \centering
    \includegraphics[width=0.8\columnwidth]{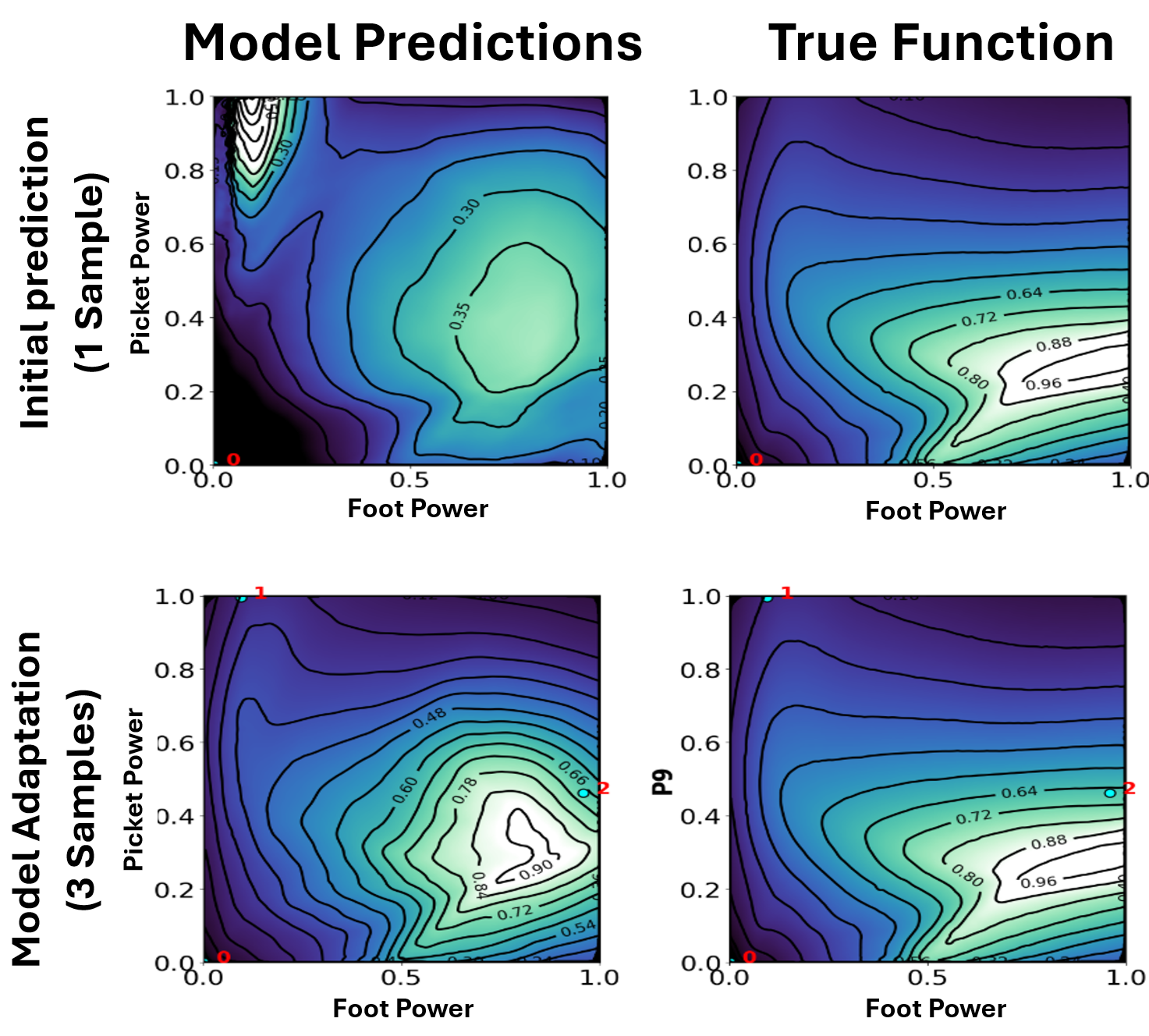}
    \caption{HL-MBO's meta-learned surrogate predictions: (top) with only one context point; (bottom) after three context points; (right) the optimization target function. We can observe that our approach achieves quick adaptation with high sample efficiency, closely approximating the true function in just three samples. This property is ideal for the limited ICF experiments possible on a shot day.}
    \label{figure:adapation}
\end{figure}

To understand the superior performance achieved by our method, we examine the predictions made by its surrogate model and assess how well it adapts. To perform this evaluation, we compare the real target function with HL-MBO predictions. For HL-MBO, first we evaluate its initial predictions when using one context point (1 sample), so the information provided to the surrogate comes only from the first evaluation sample. Moreover, we assess how HL-MBO adapts by querying it after three samples have been collected from the target function (3 Samples). Figure \ref{figure:adapation} shows the results of the evaluation. From these results, we can see that HL-MBO effectively uses information from source tasks to identify regions with a higher likelihood of finding optimal solutions. Furthermore, after incorporating just a few samples, HL-MBO accurately adjusts its predictions to match the target function closely. 

\subsection{HL-MBO vs NAP Adaptation}

\begin{figure}[H]
    \centering
    \includegraphics[width=1\columnwidth]{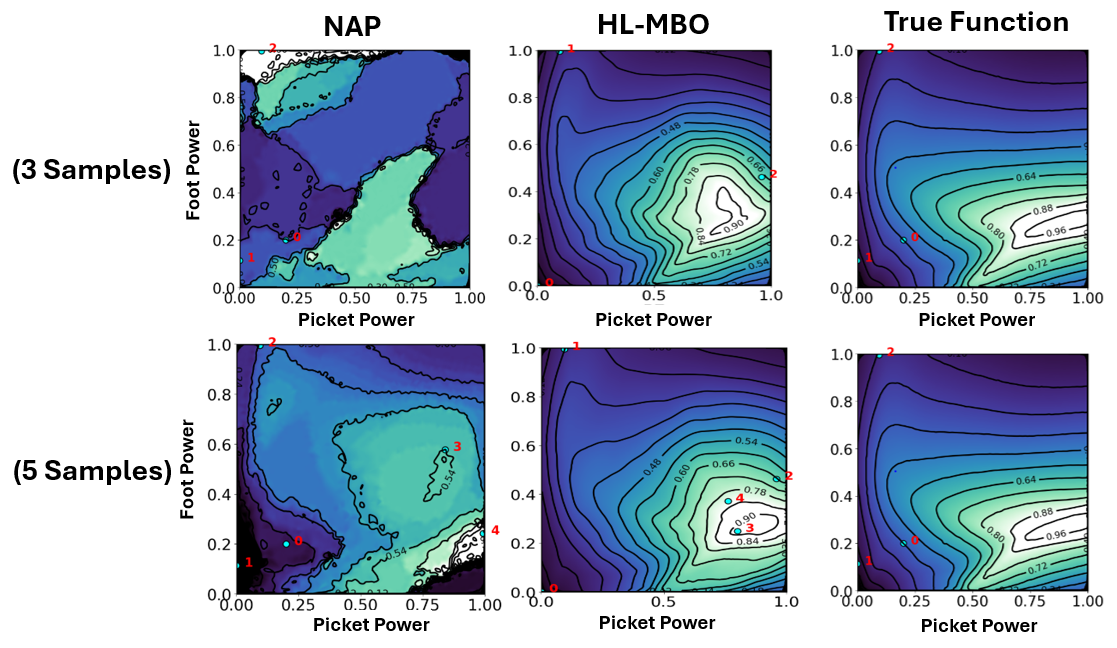}
    \caption{MBO's meta-learned surrogate prediction against NAP's surrogate prediction. MBO's predictions are more accurate, smoother and interpretable.}
    \label{figure:nap_vs_mbo}
\end{figure}

To gain deeper insight into the superior performance of HL-MBO over NAP, we compare their surrogate predictions against the state-of-the-art Meta-BO approach, which has demonstrated strong results in optimizing ICF objectives \cite{gundecha2024metalearned}. Figure \ref{figure:nap_vs_mbo} presents the outcomes of this comparison. For each trajectory, we trace the candidates proposed by the respective methods. The results show that our surrogate produces predictions that are not only more accurate and smoother but also yield surrogate outputs that are easier to interpret.

\section{No-Harm Guarantee}

A critical design requirement for human-in-the-loop optimization is robustness to imperfect expert knowledge. As shown in CoExBO \cite{adachi2024looping}, we the decay parameter $\gamma$ provides a \textbf{no-harm guarantee}: even with adversarial expert preferences, HL-MBO converges asymptotically at least as well as standard BO.

The decay mechanism in Equation 5 ensures that expert influence diminishes over time while the meta-surrogate's influence grows. The key insight lies in the behavior of $\mathscr{S}^2_\pi(x)$ as $t \to \infty$:

\begin{align}
\mathscr{S}^2_\pi(x) &= \sigma^2_\pi(x) + \gamma t^2 \sigma^2_S(x) \label{eq:decay_behavior}
\end{align}

As $t \to \infty$, since $\gamma > 0$, we have $\gamma t^2 \to \infty$, which causes $\mathscr{S}^2_\pi(x) \to \infty$.

This growth in $\mathscr{S}^2_\pi(x)$ changes the weighting in the combined mean. From Equation 2, we can derive the weights by substituting Equation 3:

\begin{align}
w_\pi(x,t) &= \frac{\sigma^2_{\mathcal{S},\pi}(x)}{\mathscr{S}^2_\pi(x)} = \frac{\mathscr{S}^2_\pi(x)\sigma^2_\mathcal{S}(x)}{\mathscr{S}^2_\pi(x)+\sigma^2_\mathcal{S}(x)} \cdot \frac{1}{\mathscr{S}^2_\pi(x)} \\
&= \frac{\sigma^2_\mathcal{S}(x)}{\mathscr{S}^2_\pi(x)+\sigma^2_\mathcal{S}(x)} \label{eq:expert_weight}
\end{align}

\begin{align}
w_S(x,t) &= \frac{\sigma^2_{\mathcal{S},\pi}(x)}{\sigma^2_\mathcal{S}(x)} = \frac{\mathscr{S}^2_\pi(x)\sigma^2_\mathcal{S}(x)}{\mathscr{S}^2_\pi(x)+\sigma^2_\mathcal{S}(x)} \cdot \frac{1}{\sigma^2_\mathcal{S}(x)} \\
&= \frac{\mathscr{S}^2_\pi(x)}{\mathscr{S}^2_\pi(x)+\sigma^2_\mathcal{S}(x)} \label{eq:surrogate_weight}
\end{align}

Note that $w_\pi(x,t) + w_S(x,t) = 1$, confirming these are proper probability weights.

\textbf{No-Harm Mechanism:} As $t \to \infty$ and $\mathscr{S}^2_\pi(x) = \sigma^2_\pi(x) + \gamma t^2 \sigma^2_S(x) \to \infty$:

\begin{align}
\lim_{t \to \infty} w_\pi(x,t) &= \lim_{t \to \infty} \frac{\sigma^2_\mathcal{S}(x)}{\mathscr{S}^2_\pi(x)+\sigma^2_\mathcal{S}(x)} \\
&= \lim_{t \to \infty} \frac{\sigma^2_\mathcal{S}(x)}{\sigma^2_\pi(x) + \gamma t^2 \sigma^2_S(x) + \sigma^2_\mathcal{S}(x)} = 0
\end{align}

\begin{align}
\lim_{t \to \infty} w_S(x,t) &= \lim_{t \to \infty} \frac{\mathscr{S}^2_\pi(x)}{\mathscr{S}^2_\pi(x)+\sigma^2_\mathcal{S}(x)} \\
&= \lim_{t \to \infty} \frac{\sigma^2_\pi(x) + \gamma t^2 \sigma^2_S(x)}{\sigma^2_\pi(x) + \gamma t^2 \sigma^2_S(x) + \sigma^2_\mathcal{S}(x)} = 1
\end{align}

Therefore, the combined mean asymptotically approaches the meta-surrogate mean:
\begin{align}
\lim_{t \to \infty} \mu_{\mathcal{S},\pi}(x) &= \lim_{t \to \infty} [w_\pi(x,t) \mu_{\pi}(x) + w_S(x,t) \mu_{\mathcal{S}}(x)] \\
&= 0 \cdot \mu_{\pi}(x) + 1 \cdot \mu_{\mathcal{S}}(x) = \mu_{\mathcal{S}}(x) \label{eq:asymptotic_mean}
\end{align}

This shows that regardless of the quality of expert preferences $\mu_\pi(x)$, the acquisition function eventually behaves identically to standard Meta-BO using only the meta-surrogate $\mathcal{S}$. The decay parameter $\gamma$ acts as a \textit{safety mechanism} that ensures:

\begin{enumerate}
\item \textbf{Early benefit:} When $t$ is small, $\gamma t^2$ is small, so $w_\pi(x,t)$ is significant, allowing expert knowledge to influence decisions
\item \textbf{Long-term safety:} When $t$ is large, $\gamma t^2$ dominates $\mathscr{S}^2_\pi(x)$, making $w_\pi(x,t) \to 0$
\item \textbf{Quadratic decay:} The quadratic form $\gamma t^2$ provides smooth but decisive transition from expert-guided to data-driven optimization
\end{enumerate}

In our setting, we use the term \textbf{no-harm} to refer to this asymptotic safety behavior of the preference contribution: the decay schedule progressively suppresses the influence of $\pi$ and causes the acquisition rule to follow the meta-surrogate more closely as optimization proceeds. This interpretation matches the role played by the decay mechanism in HL-MBO and the way it is used throughout the paper.

\section{Explainability Approaches}

\subsection{Shapley Values}

The Shapley value $\phi_j$ for feature $j$ is defined as the average marginal contribution of feature $j$ across all possible permutations of features in $x$ as:
\begin{equation*}
    \phi_j = \mathop{\sum}_{S\subseteq J \backslash\{j\}} \frac{|S|!(|J|-|S|-1)!}{|J|!}[v_{x,\mathcal{S}}(S\cup{j})-v_{x,\mathcal{S}}(S)]
\end{equation*}
where $J$ is the set of all the features, $S$ is a subset of features that does not include feature $j$ and, $v_{x,\mathcal{S}}(S)$ is a function which measures the contribution that features $S$ have to the prediction $\mathcal{S}(x)$. 

In this work, to calculate Shapley values, we utilize KernelExplainer from SHAP \cite{NIPS2017_7062}. 

\subsection{LIME}

Given an input $x \in \mathbb{R}^d$, LIME proceeds as follows:

Perturbation Sampling: Generate a set of perturbed instances $\{x'_i\}_{i=1}^N$ around $x$ by randomly modifying the features of $x$.

Prediction Querying: Evaluate the black-box model $\mathcal{S}$ on each perturbed sample to obtain predictions $\mathcal{S}(x'_i)$.

Weighting by Proximity: Assign a proximity weight $\pi_x(x'_i)$ to each perturbed instance, typically using an exponential kernel based on the distance between $x$ and $x'_i$.

Local Surrogate Modeling: Fit an interpretable model $g \in G$ (e.g., a sparse linear model) to the weighted dataset $\{(x'_i, \mathcal{S}(x'_i), \pi_x(x'_i))\}$ by minimizing the objective:
\[
\arg\min_{g \in G} \mathcal{L}(\mathcal{S}, g, \pi_x) + \Omega(g),
\]
where $\mathcal{L}$ is a loss function that captures how well $g$ approximates $\mathcal{S}$ locally, and $\Omega(g)$ is a complexity penalty that encourages interpretability.

Explanation Extraction: The coefficients of $g$ serve as feature attributions, quantifying each feature's contribution to $\mathcal{S}(x)$.

LIME is model-agnostic, requiring only black-box access to $\mathcal{S}$, and is applicable across a variety of data modalities, including tabular, textual, and image data \cite{ribeiro2016whyitrustyou}.
